\documentclass[journal]{IEEEtran}

\usepackage{amsmath,amssymb,amsfonts}
\usepackage{algorithm}
\usepackage{algorithmicx}
\usepackage{algpseudocode}

\usepackage{graphicx}
\usepackage{textcomp}
\usepackage{multirow}
\usepackage{xcolor}
\usepackage{booktabs}
\usepackage{siunitx}
\usepackage{hyperref}

\usepackage{flushend}

\title{ZKP-FedEval: Verifiable and Privacy-Preserving Federated Evaluation using Zero-Knowledge Proofs}

\author{
    Daniel~Commey,
    Benjamin~Appiah,
    Griffith~S.~Klogo,
    and~Garth~V.~Crosby%
    \thanks{D. Commey and G. V. Crosby are with the Department of Multidisciplinary Engineering, Texas A\&M University, College Station, TX 77843 USA (e-mail: dcommey@tamu.edu; gvcrosby@tamu.edu).}
    \thanks{B. Appiah is with the Department of Computer Science, Ho Technical University, Ho, Volta Region, Ghana (e-mail: bappiah@htu.edu.gh).}
    \thanks{G. S. Klogo is with the Department of Computer Engineering, Kwame Nkrumah University of Science and Technology (KNUST), Kumasi, Ghana (e-mail: gsklogo.coe@knust.edu.gh).}
}

\begin{document}
\maketitle

\begin{abstract}
Federated Learning (FL) enables collaborative model training on decentralized data without exposing raw data. However, the evaluation phase in FL may leak sensitive information through shared performance metrics. In this paper, we propose a novel protocol that incorporates Zero-Knowledge Proofs (ZKPs) to enable privacy-preserving and verifiable evaluation for FL. Instead of revealing raw loss values, clients generate a succinct proof asserting that their local loss is below a predefined threshold. Our approach is implemented without reliance on external APIs, using self-contained modules for federated learning simulation, ZKP circuit design, and experimental evaluation on both the MNIST and Human Activity Recognition (HAR) datasets. We focus on a threshold-based proof for a simple Convolutional Neural Network (CNN) model (for MNIST) and a multi-layer perceptron (MLP) model (for HAR), and evaluate the approach in terms of computational overhead, communication cost, and verifiability.
\end{abstract}


\section{Introduction}

\IEEEPARstart{F}{ederated} Learning (FL) has emerged as a practical method for training machine learning models on decentralized data, offering privacy advantages over centralized approaches \cite{mcmahan_communication-efficient_2023}. However, its distributed nature introduces significant trust and security challenges throughout its lifecycle. For instance, the training phase is vulnerable to data poisoning attacks, which requires robust defense mechanisms to ensure client integrity \cite{commey_bayesian_2025}. Orthogonal to securing training, the evaluation phase poses its own privacy challenges. Sharing even scalar metrics like loss or accuracy may enable inference attacks that leak sensitive information \cite{nasr_comprehensive_2018, melis_exploiting_2019}. Existing countermeasures such as Differential Privacy \cite{commey_securing_2024} introduce noise that degrades model utility, and secure aggregation methods \cite{bonawitz_practical_2017} do not verify the computation's integrity.

Zero-Knowledge Proofs (ZKPs) \cite{weizmann_institute_of_science_knowledge_2019} offer a promising technique to address these challenges. ZKPs enable a client (the Prover) to demonstrate that its local loss is below a threshold without revealing the precise value or underlying data. In our protocol, each client computes its local loss based on the given global model and then generates a ZKP attesting to this fact. The server verifies the proof and aggregates the simple binary outcome, preserving privacy while ensuring result correctness.

Our contributions are:
\begin{itemize}
    \item A threshold-based ZKP protocol for FL evaluation that is entirely self-contained.
    \item A detailed methodology for implementing the ZKP circuit using Circom \cite{belles-munoz_circom_2022} and Groth16 \cite{groth_size_2016}, including the integration of a CNN model for MNIST \cite{lecun_gradient-based_1998} and an MLP-style model for HAR.
    \item A prototype simulation in Python that integrates local computation, ZKP proof generation, and verification.
    \item Experimental evaluation on both MNIST and HAR datasets to assess computational overhead, communication cost, and verifiability.
\end{itemize}

The rest of the paper is organized as follows. Section~\ref{sec:related_work} discusses related work. Section~\ref{sec:preliminaries} reviews preliminaries on FL and ZKPs. Section~\ref{sec:proposed_method} details our proposed protocol and circuit design. Section~\ref{sec:system_architecture} describes the system architecture. Section~\ref{sec:experiment_setup} outlines the experimental setup. Section~\ref{sec:results_discussion} presents our findings. Section~\ref{sec:security_privacy} analyzes security and privacy guarantees. Finally, Section~\ref{sec:conclusion} concludes the paper and outlines future work.

\section{Related Work}
\label{sec:related_work}
Privacy in federated learning has received significant attention. Prior work has shown that sharing model updates or evaluation metrics can lead to privacy breaches \cite{nasr_comprehensive_2018, melis_exploiting_2019}. Techniques such as Differential Privacy \cite{dwork_algorithmic_2014} and Secure Aggregation \cite{bonawitz_practical_2017} have been proposed; however, they either degrade model performance or do not verify computation integrity.

Recent efforts have explored the application of ZKPs in machine learning \cite{ghodsi_safetynets_2017, juvekar_gazelle_2018} and federated learning \cite{xu_verifynet_2020, kalapaaking_auditable_2024}. The versatility of ZKPs allows them to secure various system layers, from proving hardware identity in blockchain-based IoT systems \cite{commey_securing_2024-1} to verifying complex computations. In contrast to prior work in FL, our approach focuses specifically on the evaluation phase, using a lightweight, threshold-based ZKP protocol implemented entirely in-house without external dependencies.

\section{Preliminaries} \label{sec:preliminaries} 
We establish notation and background for Federated Learning (FL) and Zero-Knowledge Proofs (ZKPs).

\subsection{Federated Learning (FL)} 
FL enables $N$ clients $C_i \in \{C_1, \dots, C_N\}$, each holding a private dataset $D_i = \{(x_{ij}, y_{ij})\}_{j=1}^{|D_i|}$, to collaboratively train a global model $W \in \mathbb{R}^d$. The goal is minimizing a global loss $\mathcal{L}_{global}(W)$, typically the weighted average of local losses $\mathcal{L}_i(W)$:
\begin{equation}
\min_{W \in \mathbb{R}^d} \mathcal{L}_{global}(W) = \sum_{i=1}^{N} p_i \mathcal{L}_i(W)
\end{equation}
where 
\begin{equation}
\mathcal{L}_i(W) = \frac{1}{|D_i|} \sum_{(x, y) \in D_i} \ell(W; x, y)
\end{equation}
and $p_i = |D_i| / \sum_{k} |D_k|$. $\ell$ is the sample loss function.

The standard algorithm is Federated Averaging (FedAvg) \cite{mcmahan_communication-efficient_2017}. In round $t$:
\begin{enumerate}
\item Server selects clients $S_t \subseteq \{C_1, \dots, C_N\}$ and distributes $W_t$.
\item Each client $i \in S_t$ computes $W_{t+1}^i = \text{LocalUpdate}(W_t, D_i)$ (e.g., via local SGD).
\item Clients send updates (e.g., $W_{t+1}^i$ or $\Delta_i = W_{t+1}^i - W_t$) to the server.
\item Server aggregates:
\end{enumerate}
\begin{equation}
W_{t+1} = \sum_{i \in S_t} \frac{|D_i|}{\sum_{k \in S_t} |D_k|} W_{t+1}^i
\end{equation}

FL Evaluation: Performance is assessed by having clients compute local metrics, such as loss $L_i(W_t)$ over a local test set $D_i^{test}$:
\begin{equation}
L_i(W_t) = \frac{1}{|D_i^{test}|} \sum_{(x, y) \in D_i^{test}} \ell(W_t; x, y)
\end{equation}
Reporting $L_i(W_t)$ directly can leak private information \cite{nasr_comprehensive_2018}.

\subsection{Zero-Knowledge Proofs (ZKPs)}
A ZKP \cite{weizmann_institute_of_science_knowledge_2019} allows a Prover ($\mathcal{P}$) to convince a Verifier ($\mathcal{V}$) that a statement $x$ is true, without revealing the secret witness $w$ that proves it. For an NP-language $L = \{x \mid \exists w : (x, w) \in R\}$, where $R$ is a polynomial-time verifiable relation, a ZKP system must satisfy:
\begin{itemize}
\item Completeness: If $(x, w) \in R$, an honest $\mathcal{P}$ convinces an honest $\mathcal{V}$.
\item Soundness: If $x \notin L$, no cheating $\mathcal{P}^*$ can convince $\mathcal{V}$ (except with negligible probability $\epsilon$).
\item Zero-Knowledge: $\mathcal{V}$ learns nothing beyond $x \in L$. For any PPT $\mathcal{V}^*$, a simulator $\mathcal{S}$ can generate indistinguishable transcripts given only $x$.
\end{itemize}

zk-SNARKs (Zero-Knowledge Succinct Non-Interactive Arguments of Knowledge) \cite{groth_size_2016} are ZKPs that are also:
\begin{itemize}
\item Succinct: Proofs $\pi$ are short ($|\pi| = \text{poly}(\lambda)$), and verification is fast ($T_{\mathcal{V}} = \text{poly}(\lambda, |x|)$), often independent of the witness $w$ and computation size.
\item Non-Interactive: The proof $\pi$ is a single message.
\item Argument of Knowledge: $\mathcal{P}$ must "know" $w$.
\end{itemize}

zk-SNARKs often require expressing the relation $R$ as an arithmetic circuit or Rank-1 Constraint System (R1CS) over $\mathbb{F}_p$. They typically use three algorithms: Setup $\mathcal{G}(1^\lambda, R) \rightarrow (pk, vk)$, Prove $\mathcal{P}(pk, x, w) \rightarrow \pi$, and Verify $\mathcal{V}(vk, x, \pi) \rightarrow \{\text{accept, reject}\}$. Some schemes (like Groth16) require a trusted setup for $\mathcal{G}$.

\section{Proposed Method: ZKP-based Private Evaluation}
\label{sec:proposed_method}
We propose a protocol in which clients prove that their computed local loss \(L_i\) is below a predefined threshold \(T\) without revealing its exact value. The core components are as follows:

\subsection{Protocol Overview}
The protocol proceeds in the following steps:
\begin{enumerate}
    \item Server Initialization:
    \begin{itemize}
        \item The server publishes the global model \(W_g\), loss threshold \(T\), and a round-specific nonce \(N_{\text{round}}\).
    \end{itemize}
    \item Local Computation:
    \begin{itemize}
        \item Each client computes its local loss \(L_i\) on a subset of its data (from MNIST or HAR).
    \end{itemize}
    \item Proof Generation:
    \begin{itemize}
        \item If \(L_i < T\), the client generates a ZKP using a Circom circuit that verifies the loss calculation and threshold comparison.
    \end{itemize}
    \item Verification and Aggregation:
    \begin{itemize}
        \item The server verifies each proof, aggregates the binary outcomes (valid or not), and estimates the performance of the global model.
    \end{itemize}
\end{enumerate}

\subsection{Detailed Algorithm}
Algorithm~\ref{alg:zkp_eval} details the steps for our ZKP-based evaluation protocol.

\begin{algorithm}[!ht]
\caption{ZKP-Based FL Evaluation Protocol}
\label{alg:zkp_eval}
\begin{algorithmic}[1]
\Require Global model \(W_g\), loss threshold \(T\), round nonce \(N_{\text{round}}\), client dataset \(D_i\), ZKP proving key \(pk\), verification key \(vk\)
\Ensure Aggregated evaluation result \(R\)
\State Server: Publish \((W_g, T, N_{\text{round}})\) to all clients.
\ForAll{client \(i\) in the selected group}
    \State Client \(i\) computes local loss: \(L_i = \frac{1}{|D_i|} \sum_{x \in D_i} \mathcal{L}(W_g, x)\)
    \If{\(L_i < T\)}
        \State Generate ZKP: \((\pi_i, pub\_input_i) \gets \textsc{GenerateZKP}(pk, W_g, D_i, T, N_{\text{round}})\)
        \State Send \((\pi_i, pub\_input_i)\) to the server.
    \Else
        \State Send failure indicator (or no proof) to the server.
    \EndIf
\EndFor
\State Server: For each received proof, execute:
\State \quad \((b_i \gets \textsc{VerifyZKP}(vk, pub\_input_i, \pi_i))\)
\State \quad if \(b_i = \text{true}\) then count client \(i\) as valid.
\State Aggregate valid outcomes into result \(R\) (e.g., count valid proofs).
\State \Return \(R\)
\medskip
\Procedure{GenerateZKP}{$pk, W_g, D_i, T, N_{\text{round}}$}
    \State Compute \(H_{Wg} \gets \text{hash}(W_g)\)
    \State Set public inputs: \(pub\_input \gets \{H_{Wg}, T, N_{\text{round}}\}\)
    \State Represent model weights and data in fixed-point format.
    \State Compute the witness using the CNN/MLP forward pass and loss function.
    \State Generate proof \(\pi\) using snarkjs \cite{baylina_iden3snarkjs_2020} (or equivalent) with \(pk\), \(pub\_input\), and the witness.
    \State \Return \((\pi, pub\_input)\)
\EndProcedure
\medskip
\Procedure{VerifyZKP}{$vk, pub\_input, \pi$}
    \State Verify that \(pub\_input\) contains the correct \(T\), \(N_{\text{round}}\), and hash \(H_{Wg}\).
    \State Use the ZKP verification routine to check \(\pi\) with \(vk\) and \(pub\_input\).
    \State \Return verification result (true/false).
\EndProcedure
\end{algorithmic}
\end{algorithm}

\subsection{ZKP Circuit Overview}
The ZKP circuit, written in Circom \cite{belles-munoz_circom_2022}, performs the following:
\begin{itemize}
    \item Verifies that the private input weights \(W_g\) (in fixed-point representation) hash to the public hash \(H_{Wg}\).
    \item Simulates the forward pass of the chosen model (either the CNN or MLP), computing the loss for a given input sample.
    \item Uses fixed-point arithmetic to compute the loss and enforces that the averaged loss is below \(T\).
    \item Includes the round nonce \(N_{\text{round}}\) in the computation to prevent replay attacks.
\end{itemize}

\subsection{Instantiation with Groth16}
We use the Groth16 zk-SNARK scheme \cite{groth_size_2016}:
\begin{itemize}
    \item Trusted Setup: Generate the proving key \(pk\) and verification key \(vk\).
    \item Proof Generation: Each client computes a witness and generates a proof \(\pi\) using the circuit.
    \item Proof Verification: The server verifies the proofs using \(vk\).
\end{itemize}

\section{System Architecture}
\label{sec:system_architecture}

\begin{figure}[ht]
    \centering
    \includegraphics[width=\linewidth]{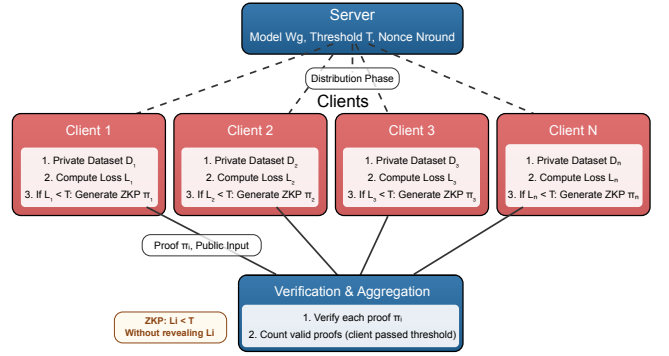}
    \caption{ZKP-FedEval System Architecture: The server distributes the model, threshold, and nonce. Clients compute local loss and generate ZKPs if loss is below threshold. The server verifies proofs and aggregates binary outcomes.}
    \label{fig:system_architecture}
\end{figure}

The ZKP-FedEval system consists of three primary components working together to enable privacy-preserving federated evaluation. Figure~\ref{fig:system_architecture} illustrates this architecture, showing how the server, clients, and ZKP infrastructure interact.

\subsection{Core Components}
The server functions as the central coordinator, broadcasting the global model $W_g$ to clients, defining the loss threshold $T$, and generating unique round-specific nonces to prevent replay attacks. After receiving proofs, it verifies them and aggregates the binary outcomes to assess global model performance.

Clients operate on their private datasets, computing local loss values and checking if they fall below the threshold. When this condition is met, they generate zero-knowledge proofs attesting to this fact without revealing the actual loss values.

The ZKP infrastructure provides the technical foundation with Circom templates defining verification logic, Groth16 protocol implementation, key management services, and fixed-point arithmetic components for handling floating-point operations within the constraints of the ZKP circuits.

\section{Experiment Setup}
\label{sec:experiment_setup}
Our experiments evaluated ZKP-FedEval using two datasets: MNIST \cite{lecun_gradient-based_1998} (28$\times$28 grayscale images of handwritten digits with 10 classes) and UCI HAR \cite{anguita_public_2013} (Human Activity Recognition data from smartphone sensors with 6 activities). For MNIST, we implemented a 2-layer CNN with channels evolving from 32 to 64, followed by 2 fully connected layers. For HAR, we used a 1D CNN with temporal convolutions (9 input channels → 64 → 128).

The experimental configuration varied the number of clients (5, 10, 15, 20) and loss thresholds (0.5, 1.0, 1.5) with a fixed batch size of 32. Server training consisted of 1 epoch on 10\% of training data, while client evaluation used test set data evenly distributed across clients. We ran each configuration with 3 random seeds and tested both IID and basic non-IID data distribution variants \cite{hsu_measuring_2019}.

Our ZKP implementation used a simple loss threshold comparison circuit with the Groth16 protocol \cite{groth_size_2016}, 6 decimal places of fixed-point precision, and SHA-256 for model hash and nonce. All proof generation and verification ran on CPU, with parallel client evaluation simulation but sequential proof verification at the server.

Table~\ref{tab:models} describes the architectures of the two neural network models in greater detail.

\begin{table}[!ht]
\centering
\caption{Model Architectures and Descriptions}
\label{tab:models}
\resizebox{\linewidth}{!}{%
\begin{tabular}{@{}lll@{}}
\toprule
\textbf{Model} & \textbf{Architecture} & \textbf{Dataset} \\ \midrule
\texttt{MNIST\_CNN} & \begin{tabular}[c]{@{}l@{}}Conv2D(1, 16, kernel=5, padding=2) $\rightarrow$ ReLU $\rightarrow$ MaxPool2D(2) \\ Conv2D(16, 32, kernel=5, padding=2) $\rightarrow$ ReLU $\rightarrow$ MaxPool2D(2) \\ Dropout(0.5) $\rightarrow$ FC(32$\times$7$\times$7, 10)\end{tabular} & MNIST \\ \midrule
\texttt{HARCNN} & \begin{tabular}[c]{@{}l@{}}Fully Connected (input\_features $\rightarrow$ 128) $\rightarrow$ ReLU $\rightarrow$ Dropout(0.5) \\ Fully Connected (128 $\rightarrow$ 64) $\rightarrow$ ReLU $\rightarrow$ Dropout(0.3) \\ Fully Connected (64 $\rightarrow$ 6)\end{tabular} & HAR \\ \bottomrule
\end{tabular}%
}
\end{table}

\section{Results and Discussion}
\label{sec:results_discussion}

This section presents an analysis of the performance of our ZKP-based federated evaluation protocol, focusing on computation time, communication costs, and the effectiveness of the loss threshold mechanism.

\subsection{Computation Time}

\begin{figure}[ht]
    \centering
    \includegraphics[width=\linewidth]{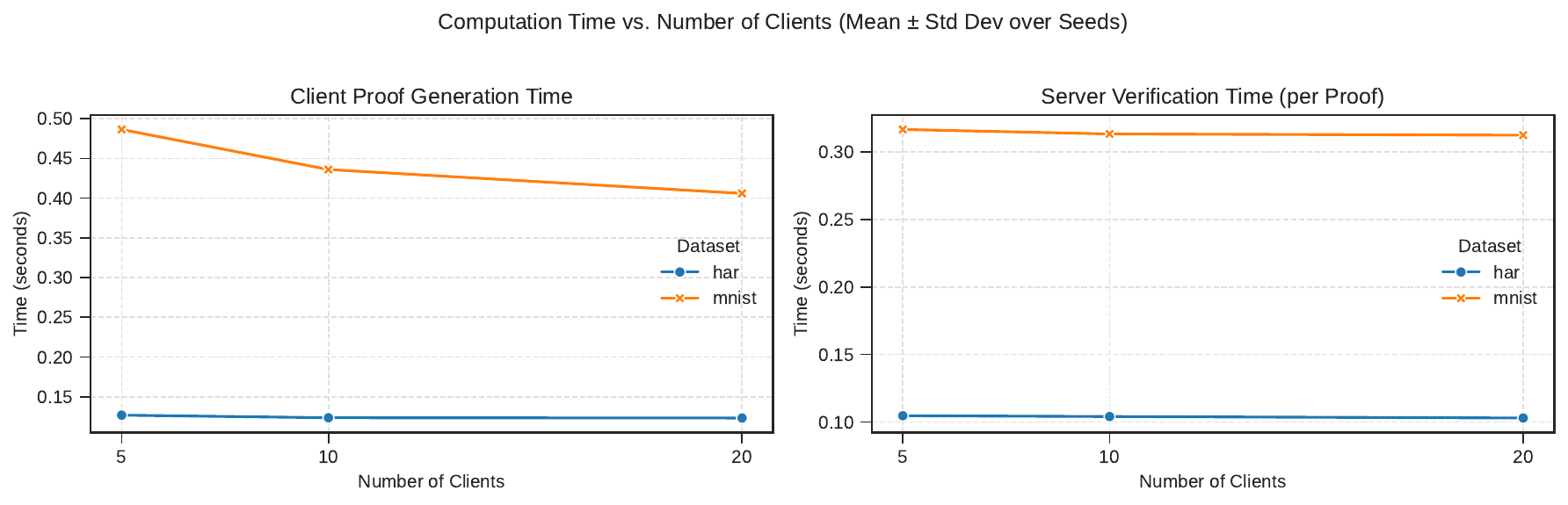}
    \caption{Average client-side proof generation time and server-side verification time as a function of the number of clients.}
    \label{fig:time_vs_clients}
\end{figure}

Figure~\ref{fig:time_vs_clients} shows the client-side proof generation time and the server-side verification time as a function of the number of clients. Key observations include:
\begin{itemize}
    \item Client Proof Generation Time: The average time required by clients to evaluate the model and generate a proof (when the loss is below the threshold) remains stable or slightly decreases as the number of clients increases. This behavior is expected since a fixed data percentage distributed among a larger client set implies that each client processes a smaller volume of data.
    \item For the MNIST dataset (using the CNN model), the average proof generation time is approximately 0.4--0.5\,s, while for the HAR dataset (using the MLP model) it is around 0.12--0.13\,s. This difference is largely attributed to the additional computational load of evaluating a CNN compared to an MLP.
    \item Server Verification Time: The average time to verify a proof is essentially constant across different numbers of clients (approximately 0.31--0.32\,s for MNIST and about 0.10\,s for HAR, for valid proofs). Note that total server verification time scales linearly with the number of valid proofs received.
\end{itemize}

\subsection{Communication Cost}

\begin{figure}[ht]
    \centering
    \includegraphics[width=\linewidth]{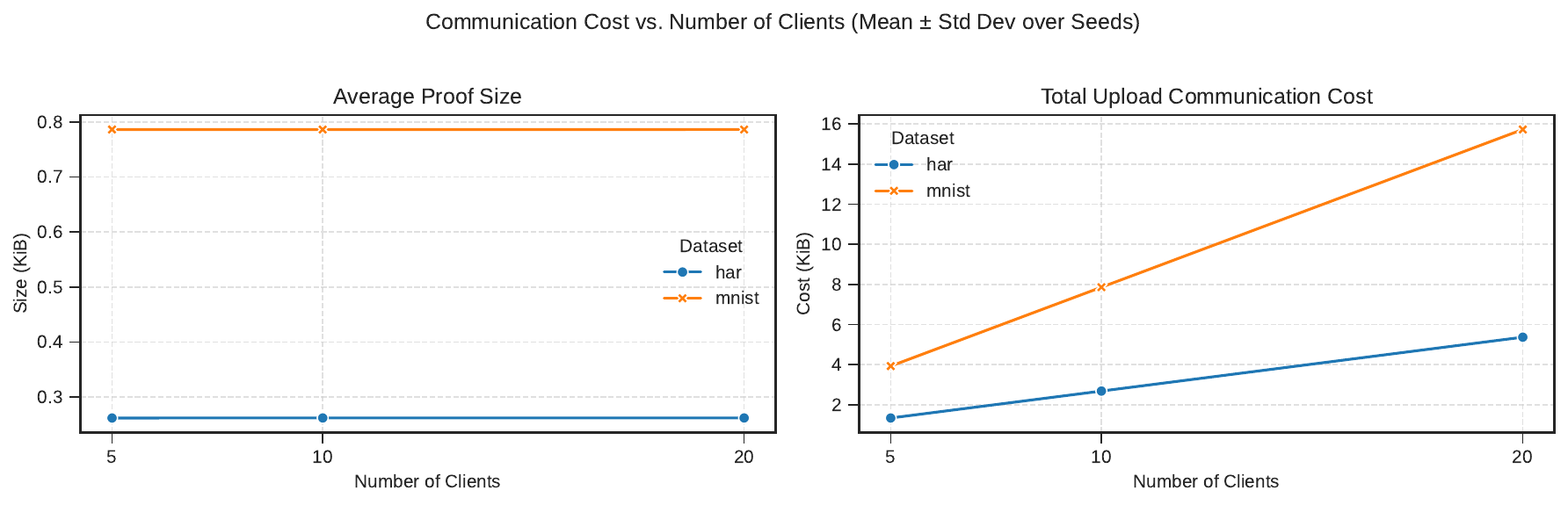}
    \caption{Communication cost (total upload size) as a function of the number of clients.}
    \label{fig:comm_cost_vs_clients}
\end{figure}

Figure~\ref{fig:comm_cost_vs_clients} presents the communication overhead associated with the proof exchange process.
\begin{itemize}
    \item The average proof size remains constant at approximately 0.79\,KiB for the MNIST dataset and around 0.26\,KiB for HAR (when proofs are generated), which is consistent with the fixed proof size characteristic of Groth16 proofs \cite{groth_size_2016}.
    \item The total communication cost scales linearly with the number of clients submitting valid proofs. For the MNIST dataset—where the validation rate was 100\%—the total cost increases directly with the client count, while for HAR the cost is lower due to a reduced validation rate.
\end{itemize}

\subsection{Validation Rate and Threshold Effectiveness}

\begin{figure}[ht]
    \centering
    \includegraphics[width=0.7\linewidth]{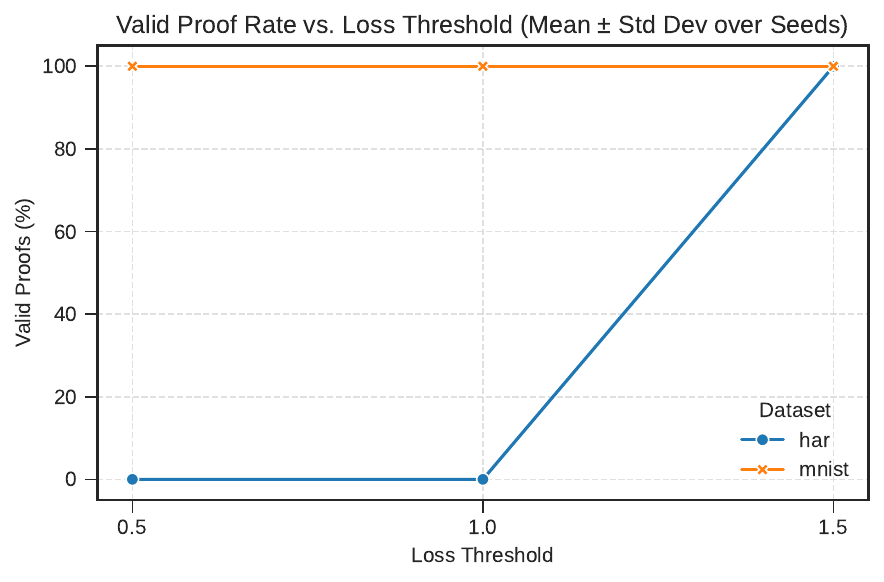}
    \caption{Validation rate (percentage of clients generating valid proofs) for different loss thresholds.}
    \label{fig:validation_vs_threshold}
\end{figure}

Figure~\ref{fig:validation_vs_threshold} and the summary table (Table~\ref{tab:performance_summary}) illustrate the impact of the loss threshold on the validation rate:
\begin{itemize}
    \item Threshold Impact: As the threshold value increases, more clients are able to generate valid proofs, thereby increasing the validation rate. This behavior aligns with theoretical expectations and previous findings on threshold-based approaches in federated settings \cite{hsu_measuring_2019}.
    
    \item For the MNIST dataset, the validation rate remained at 100\% across all tested thresholds (0.5, 1.0, 1.5), indicating that the trained CNN model consistently achieved a loss below even the lowest threshold. This high success rate is consistent with the relatively high accuracy typically achieved by CNN models on the MNIST dataset \cite{lecun_gradient-based_1998}.
    
    \item In contrast, the HAR dataset exhibited a lower validation rate (averaging around 33\%) for the lowest threshold (0.5), while higher thresholds (1.0, 1.5) resulted in nearly 100\% validation. This suggests that the MLP model for HAR often had losses above the strict threshold but satisfied higher thresholds, reflecting the greater inherent difficulty in accurately classifying human activity data compared to handwritten digit recognition \cite{kairouz_advances_2021}.
    
    \item The differential response to thresholds between datasets demonstrates the importance of carefully calibrating threshold values based on expected model performance, data complexity, and application requirements.
\end{itemize}

Figure~\ref{fig:validation_vs_threshold} shows how varying the threshold affects both the validation rate and the information revealed to the server. A higher threshold increases the percentage of clients that can generate valid proofs but potentially reduces the meaningfulness of the binary signal received by the server. Conversely, a lower threshold provides a more stringent quality guarantee but may exclude clients with reasonably good performance. This trade-off between inclusivity and meaningfulness must be carefully balanced in real-world deployments.

\subsection{Scalability Analysis}

\begin{figure}[ht]
    \centering
    \includegraphics[width=0.7\linewidth]{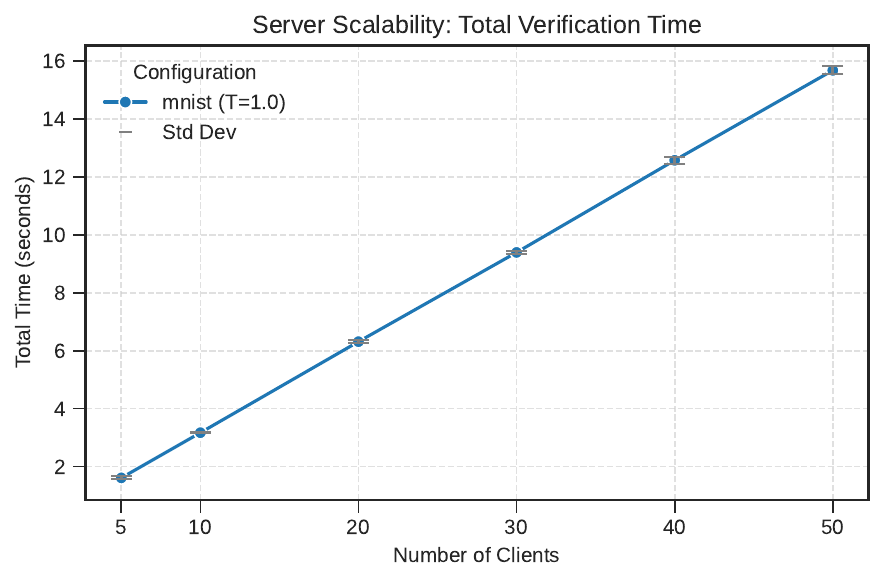}
    \caption{Total server verification time as the number of clients increases, demonstrating linear scaling. Results shown for the MNIST dataset with threshold T=1.0.}
    \label{fig:scalability}
\end{figure}

To assess the scalability of our approach, we performed additional experiments increasing the number of clients up to 50 while measuring total server-side verification time. As shown in Figure~\ref{fig:scalability}, the relationship remains linear, confirming that our approach can scale to larger FL deployments with predictable overhead.

The server can easily parallelize verification operations, as each proof is verified independently. With appropriate hardware, the server's verification capacity can be expanded horizontally to accommodate larger client populations.

For client-side operations, we observed that proof generation time is primarily determined by the model complexity rather than dataset size. The CNN model for MNIST required approximately 3-4 times more computation time than the MLP model for HAR, despite similar proof sizes. This suggests that for more complex models like deep CNNs or modern architectures (e.g., ResNets, DenseNets), the circuit design and witness generation process would need significant optimization \cite{xu_verifynet_2020}.

\subsection{Summary Table}

Table~\ref{tab:performance_summary} provides a comprehensive overview of the measured metrics, including average proof generation time, verification time, communication cost per client, and overall validation rate for both datasets.

\begin{table}[ht]
\centering
\caption{Summary of Performance Metrics (Mean over Seeds)}
\label{tab:performance_summary}
\resizebox{\linewidth}{!}{%
\begin{tabular}{llrrrr}
\toprule
 &  & Client Time (s) & Verify Time (s) & Comm Cost (KiB) & Valid Proofs (\%) \\
Dataset & Num Clients &  &  &  &  \\
\midrule
\multirow[t]{3}{*}{har} & 5 & 0.13 & 0.10 & 0.26 & 33.33 \\
 & 10 & 0.12 & 0.10 & 0.26 & 33.33 \\
 & 20 & 0.12 & 0.10 & 0.26 & 33.33 \\
\cline{1-6}
\multirow[t]{3}{*}{mnist} & 5 & 0.49 & 0.32 & 0.79 & 100.00 \\
 & 10 & 0.44 & 0.31 & 0.79 & 100.00 \\
 & 20 & 0.41 & 0.31 & 0.79 & 100.00 \\
\cline{1-6}
\bottomrule
\end{tabular}%
}
\end{table}

\subsection{Discussion and Limitations}

The experimental results validate that:
\begin{itemize}
    \item Server verification time and proof size are constant per proof, confirming a key advantage of zk-SNARKs over other cryptographic approaches \cite{groth_size_2016}.
    
    \item Total server verification time and total communication cost increase linearly with the number of clients that submit valid proofs, indicating good scalability properties.
    
    \item Client-side proof generation time is primarily influenced by the computational cost of local evaluation, and it is notably higher for complex models (e.g., CNN on MNIST) compared to simpler models (e.g., MLP on HAR). This suggests that proof generation overhead could become significant for deeper neural network architectures \cite{xu_verifynet_2020}.
    
    \item The chosen loss threshold effectively filters clients, but its precise value must be tuned based on the application, model complexity, and data distribution \cite{hsu_measuring_2019}.
\end{itemize}

Our approach shows promise, but several limitations remain:

\begin{itemize}
    \item Circuit Complexity: The current circuit only verifies the threshold condition rather than the full forward pass and loss computation. A complete implementation would require significant circuit engineering to represent all neural network operations efficiently \cite{ghodsi_safetynets_2017}.
    
    \item Trusted Setup: The Groth16 protocol requires a trusted setup phase, which could present practical challenges in real-world deployments \cite{groth_size_2016}. More recent ZKP systems with transparent setups (e.g., PLONK, STARKs) could address this issue but typically have larger proof sizes or verification times.
    
    \item Binary Information: Our protocol only reveals whether a client's loss is below the threshold, providing minimal information about actual model performance. This could limit its utility for fine-grained evaluation and debugging purposes \cite{kairouz_advances_2021}.
    
    \item Computational Overhead: While acceptable for the models tested, the proof generation overhead could become prohibitive for larger neural networks or resource-constrained client devices \cite{bonawitz_practical_2017}.
    
    \item Data and Model Privacy: Although our approach protects evaluation metrics, it does not address other privacy concerns in FL, such as information leakage through model updates \cite{zhu_deep_2019}.
\end{itemize}

Despite these limitations, our ZKP-based approach represents a significant step toward enhancing both privacy and integrity in federated evaluation scenarios where sharing raw metrics might be problematic.

\section{Security and Privacy Analysis}
\label{sec:security_privacy}
The security guarantees stem from the core properties of zk-SNARKs \cite{weizmann_institute_of_science_knowledge_2019}:
\begin{itemize}
    \item Zero-Knowledge: The server learns only that a client's loss is below \(T\) without obtaining its value.
    \item Soundness: A client cannot generate a valid proof unless the computed loss meets the threshold condition.
\end{itemize}
Inclusion of a round-specific nonce \(N_{\text{round}}\) prevents replay attacks. While our approach assumes a trusted setup for Groth16 \cite{groth_size_2016}, future work may explore transparent setup alternatives.

The privacy guarantees of our approach are stronger than typical FL evaluation protocols, where clients share raw loss values that might leak information about their dataset \cite{melis_exploiting_2019}. By revealing only a single bit of information (whether the loss is below the threshold), we significantly reduce the risk of inference attacks that attempt to deduce properties of client data from evaluation metrics \cite{nasr_comprehensive_2018}.

From an integrity standpoint, the soundness property ensures that malicious clients cannot falsely claim their loss is below the threshold, providing stronger verification guarantees than standard FL systems that rely on trust or statistical methods to detect anomalous behavior \cite{kairouz_advances_2021}.

\section{Conclusion}
\label{sec:conclusion}
We have presented ZKP-FedEval, a practical protocol for verifiable, privacy-preserving federated evaluation using Zero-Knowledge Proofs. By enabling clients to prove that their local loss is below a set threshold without revealing the raw metric, our approach enhances trust in FL evaluation \cite{kairouz_advances_2021}. The prototype—implemented in Python with self-contained modules for model evaluation and ZKP proof generation—has been evaluated on both MNIST \cite{lecun_gradient-based_1998} and HAR datasets.

Our results indicate that the protocol incurs reasonable computational and communication overhead while providing strong privacy guarantees. For the MNIST dataset, client proof generation takes approximately 0.4-0.5 seconds, with verification requiring about 0.3 seconds. For the simpler HAR dataset, these times are even lower.

Future work will focus on optimizing the ZKP circuit \cite{belles-munoz_circom_2022}, exploring alternative ZKP schemes with transparent setups \cite{grassi_poseidon_2021}, and extending the protocol to include richer evaluation metrics. Additionally, we aim to investigate the use of more sophisticated threshold mechanisms that could adapt to model performance over time, potentially incorporating techniques from previous work on secure aggregation \cite{bonawitz_practical_2017} and differentially private mechanisms \cite{dwork_algorithmic_2014}.

\bibliographystyle{IEEEtran}
\bibliography{references}

\begin{thebibliography}{10}
\providecommand{\url}[1]{#1}
\csname url@samestyle\endcsname
\providecommand{\newblock}{\relax}
\providecommand{\bibinfo}[2]{#2}
\providecommand{\BIBentrySTDinterwordspacing}{\spaceskip=0pt\relax}
\providecommand{\BIBentryALTinterwordstretchfactor}{4}
\providecommand{\BIBentryALTinterwordspacing}{\spaceskip=\fontdimen2\font plus
\BIBentryALTinterwordstretchfactor\fontdimen3\font minus \fontdimen4\font\relax}
\providecommand{\BIBforeignlanguage}[2]{{%
\expandafter\ifx\csname l@#1\endcsname\relax
\typeout{** WARNING: IEEEtran.bst: No hyphenation pattern has been}%
\typeout{** loaded for the language `#1'. Using the pattern for}%
\typeout{** the default language instead.}%
\else
\language=\csname l@#1\endcsname
\fi
#2}}
\providecommand{\BIBdecl}{\relax}
\BIBdecl

\bibitem{mcmahan_communication-efficient_2023}
\BIBentryALTinterwordspacing
H.~B. McMahan, E.~Moore, D.~Ramage, S.~Hampson, and B.~A.~y. Arcas, ``Communication-{Efficient} {Learning} of {Deep} {Networks} from {Decentralized} {Data},'' Jan. 2023, arXiv:1602.05629 [cs]. [Online]. Available: \url{http://arxiv.org/abs/1602.05629}
\BIBentrySTDinterwordspacing

\bibitem{commey_bayesian_2025}
\BIBentryALTinterwordspacing
D.~Commey, R.~A. Sarpong, G.~S. Klogo, W.~Bagyl-Bac, and G.~V. Crosby, ``A {Bayesian} {Incentive} {Mechanism} for {Poison}-{Resilient} {Federated} {Learning},'' Jul. 2025, arXiv:2507.12439 [cs]. [Online]. Available: \url{http://arxiv.org/abs/2507.12439}
\BIBentrySTDinterwordspacing

\bibitem{nasr_comprehensive_2018}
\BIBentryALTinterwordspacing
M.~Nasr, R.~Shokri, and A.~Houmansadr, ``\BIBforeignlanguage{en}{Comprehensive {Privacy} {Analysis} of {Deep} {Learning}: {Passive} and {Active} {White}-box {Inference} {Attacks} against {Centralized} and {Federated} {Learning}},'' Dec. 2018. [Online]. Available: \url{https://arxiv.org/abs/1812.00910v2}
\BIBentrySTDinterwordspacing

\bibitem{melis_exploiting_2019}
\BIBentryALTinterwordspacing
L.~Melis, C.~Song, E.~De~Cristofaro, and V.~Shmatikov, ``Exploiting {Unintended} {Feature} {Leakage} in {Collaborative} {Learning},'' in \emph{2019 {IEEE} {Symposium} on {Security} and {Privacy} ({SP})}, May 2019, pp. 691--706, iSSN: 2375-1207. [Online]. Available: \url{https://ieeexplore.ieee.org/document/8835269}
\BIBentrySTDinterwordspacing

\bibitem{commey_securing_2024}
\BIBentryALTinterwordspacing
D.~Commey, S.~Hounsinou, and G.~V. Crosby, ``Securing {Health} {Data} on the {Blockchain}: {A} {Differential} {Privacy} and {Federated} {Learning} {Framework},'' May 2024, arXiv:2405.11580 [cs]. [Online]. Available: \url{http://arxiv.org/abs/2405.11580}
\BIBentrySTDinterwordspacing

\bibitem{bonawitz_practical_2017}
\BIBentryALTinterwordspacing
K.~Bonawitz, V.~Ivanov, B.~Kreuter, A.~Marcedone, H.~B. McMahan, S.~Patel, D.~Ramage, A.~Segal, and K.~Seth, ``\BIBforeignlanguage{en}{Practical {Secure} {Aggregation} for {Privacy}-{Preserving} {Machine} {Learning}},'' in \emph{\BIBforeignlanguage{en}{Proceedings of the 2017 {ACM} {SIGSAC} {Conference} on {Computer} and {Communications} {Security}}}.\hskip 1em plus 0.5em minus 0.4em\relax Dallas Texas USA: ACM, Oct. 2017, pp. 1175--1191. [Online]. Available: \url{https://dl.acm.org/doi/10.1145/3133956.3133982}
\BIBentrySTDinterwordspacing

\bibitem{weizmann_institute_of_science_knowledge_2019}
\BIBentryALTinterwordspacing
{MIT}, S.~Goldwasser, S.~Micali, {MIT}, C.~Rackoff, and {University of Toronto}, ``The knowledge complexity of interactive proof-systems,'' in \emph{Providing {Sound} {Foundations} for {Cryptography}: {On} the {Work} of {Shafi} {Goldwasser} and {Silvio} {Micali}}, {Weizmann Institute of Science} and O.~Goldreich, Eds.\hskip 1em plus 0.5em minus 0.4em\relax Association for Computing Machinery, Oct. 2019. [Online]. Available: \url{https://dl.acm.org/citation.cfm?id=3335750}
\BIBentrySTDinterwordspacing

\bibitem{belles-munoz_circom_2022}
\BIBentryALTinterwordspacing
M.~Bellés-Muñoz, M.~Isabel, J.~L. Muñoz-Tapia, A.~Rubio, and J.~Baylina, ``Circom: {A} circuit description language for building zero-knowledge applications,'' \emph{IEEE Transactions on Dependable and Secure Computing}, vol.~20, no.~6, pp. 4733--4751, 2022, publisher: IEEE. [Online]. Available: \url{https://ieeexplore.ieee.org/abstract/document/10002421/}
\BIBentrySTDinterwordspacing

\bibitem{groth_size_2016}
J.~Groth, ``\BIBforeignlanguage{en}{On the {Size} of {Pairing}-{Based} {Non}-interactive {Arguments}},'' in \emph{\BIBforeignlanguage{en}{Advances in {Cryptology} – {EUROCRYPT} 2016}}, M.~Fischlin and J.-S. Coron, Eds.\hskip 1em plus 0.5em minus 0.4em\relax Berlin, Heidelberg: Springer, 2016, pp. 305--326.

\bibitem{lecun_gradient-based_1998}
\BIBentryALTinterwordspacing
Y.~LeCun, L.~Bottou, Y.~Bengio, and P.~Haffner, ``Gradient-based learning applied to document recognition,'' \emph{Proceedings of the IEEE}, vol.~86, no.~11, pp. 2278--2324, 1998, publisher: Ieee. [Online]. Available: \url{https://ieeexplore.ieee.org/abstract/document/726791/}
\BIBentrySTDinterwordspacing

\bibitem{dwork_algorithmic_2014}
\BIBentryALTinterwordspacing
C.~Dwork and A.~Roth, ``The algorithmic foundations of differential privacy,'' \emph{Foundations and Trends® in Theoretical Computer Science}, vol.~9, no. 3–4, pp. 211--407, 2014, publisher: Now Publishers, Inc. [Online]. Available: \url{https://www.nowpublishers.com/article/Details/TCS-042}
\BIBentrySTDinterwordspacing

\bibitem{ghodsi_safetynets_2017}
\BIBentryALTinterwordspacing
Z.~Ghodsi, {View Profile}, T.~Gu, {View Profile}, S.~Garg, and {View Profile}, ``{SafetyNets},'' \emph{Proceedings of the 31st International Conference on Neural Information Processing Systems}, pp. 4675--4684, Dec. 2017. [Online]. Available: \url{https://dlnext.acm.org/doi/abs/10.5555/3294996.3295220}
\BIBentrySTDinterwordspacing

\bibitem{juvekar_gazelle_2018}
\BIBentryALTinterwordspacing
C.~Juvekar, V.~Vaikuntanathan, and A.~Chandrakasan, ``\BIBforeignlanguage{en}{\{{GAZELLE}\}: {A} {Low} {Latency} {Framework} for {Secure} {Neural} {Network} {Inference}},'' 2018, pp. 1651--1669. [Online]. Available: \url{https://www.usenix.org/conference/usenixsecurity18/presentation/juvekar}
\BIBentrySTDinterwordspacing

\bibitem{xu_verifynet_2020}
\BIBentryALTinterwordspacing
G.~Xu, H.~Li, S.~Liu, K.~Yang, and X.~Lin, ``{VerifyNet}: {Secure} and {Verifiable} {Federated} {Learning},'' \emph{IEEE Transactions on Information Forensics and Security}, vol.~15, pp. 911--926, 2020. [Online]. Available: \url{https://ieeexplore.ieee.org/abstract/document/8765347}
\BIBentrySTDinterwordspacing

\bibitem{kalapaaking_auditable_2024}
\BIBentryALTinterwordspacing
A.~P. Kalapaaking, I.~Khalil, X.~Yi, K.-Y. Lam, G.-B. Huang, and N.~Wang, ``Auditable and verifiable federated learning based on blockchain-enabled decentralization,'' \emph{IEEE Transactions on Neural Networks and Learning Systems}, 2024, publisher: IEEE. [Online]. Available: \url{https://ieeexplore.ieee.org/abstract/document/10557507/}
\BIBentrySTDinterwordspacing

\bibitem{commey_securing_2024-1}
\BIBentryALTinterwordspacing
D.~Commey, S.~Hounsinou, and G.~V. Crosby, ``Securing {Blockchain}-based {IoT} {Systems} with {Physical} {Unclonable} {Functions} and {Zero}-{Knowledge} {Proofs},'' in \emph{2024 {IEEE} 49th {Conference} on {Local} {Computer} {Networks} ({LCN})}, Oct. 2024, pp. 1--7, iSSN: 2832-1421. [Online]. Available: \url{https://ieeexplore.ieee.org/document/10639679}
\BIBentrySTDinterwordspacing

\bibitem{mcmahan_communication-efficient_2017}
\BIBentryALTinterwordspacing
B.~McMahan, E.~Moore, D.~Ramage, S.~Hampson, and B.~A. y~Arcas, ``Communication-efficient learning of deep networks from decentralized data,'' in \emph{Artificial intelligence and statistics}.\hskip 1em plus 0.5em minus 0.4em\relax PMLR, 2017, pp. 1273--1282. [Online]. Available: \url{https://proceedings.mlr.press/v54/mcmahan17a?ref=https://githubhelp.com}
\BIBentrySTDinterwordspacing

\bibitem{baylina_iden3snarkjs_2020}
J.~Baylina, ``iden3/snarkjs,'' 2020.

\bibitem{anguita_public_2013}
\BIBentryALTinterwordspacing
D.~Anguita, A.~Ghio, L.~Oneto, X.~Parra, and J.~L. Reyes-Ortiz, ``A public domain dataset for human activity recognition using smartphones.'' in \emph{Esann}, vol.~3, 2013, pp. 3--4, issue: 1. [Online]. Available: \url{https://www.esann.org/sites/default/files/proceedings/legacy/es2013-84.pdf}
\BIBentrySTDinterwordspacing

\bibitem{hsu_measuring_2019}
\BIBentryALTinterwordspacing
T.-M.~H. Hsu, H.~Qi, and M.~Brown, ``Measuring the {Effects} of {Non}-{Identical} {Data} {Distribution} for {Federated} {Visual} {Classification},'' Sep. 2019, arXiv:1909.06335 [cs]. [Online]. Available: \url{http://arxiv.org/abs/1909.06335}
\BIBentrySTDinterwordspacing

\bibitem{kairouz_advances_2021}
\BIBentryALTinterwordspacing
P.~Kairouz, H.~B. McMahan, B.~Avent, A.~Bellet, M.~Bennis, A.~N. Bhagoji, K.~Bonawitz, Z.~Charles, G.~Cormode, and R.~Cummings, ``Advances and open problems in federated learning,'' \emph{Foundations and trends® in machine learning}, vol.~14, no. 1–2, pp. 1--210, 2021, publisher: Now Publishers, Inc. [Online]. Available: \url{https://www.nowpublishers.com/article/Details/MAL-083}
\BIBentrySTDinterwordspacing

\bibitem{zhu_deep_2019}
\BIBentryALTinterwordspacing
L.~Zhu, Z.~Liu, and S.~Han, ``Deep leakage from gradients,'' \emph{Advances in neural information processing systems}, vol.~32, 2019. [Online]. Available: \url{https://proceedings.neurips.cc/paper/2019/hash/60a6c4002cc7b29142def8871531281a-Abstract.html}
\BIBentrySTDinterwordspacing

\bibitem{grassi_poseidon_2021}
\BIBentryALTinterwordspacing
L.~Grassi, D.~Khovratovich, C.~Rechberger, A.~Roy, and M.~Schofnegger, ``Poseidon: {A} new hash function for \{{Zero}-{Knowledge}\} proof systems,'' in \emph{30th {USENIX} {Security} {Symposium} ({USENIX} {Security} 21)}, 2021, pp. 519--535. [Online]. Available: \url{https://www.usenix.org/conference/usenixsecurity21/presentation/grassi}
\BIBentrySTDinterwordspacing

\end{thebibliography}

\end{document}